\pdfoutput=1
\documentclass[11pt,a4paper]{article}
\usepackage[hyperref]{acl2020}
\usepackage{times}
\usepackage{latexsym}

\usepackage{url}
\usepackage{microtype}
\usepackage{amsmath}
\usepackage{amsfonts}
\usepackage{algorithm}
\usepackage{algpseudocode}
\usepackage{booktabs}
\usepackage{xparse}
\usepackage{graphicx}
\usepackage{multicol}
\usepackage{multirow}
\aclfinalcopy 
\usepackage{tikz}
\usetikzlibrary{shapes,arrows}
\usepackage{natbib}
\usepackage{caption}
\usepackage{subcaption}
\usepackage{tikz-dependency}
\usepackage{wrapfig}
\usepackage{pgfplots}
\usepackage{filecontents}
\usepackage[T1]{fontenc}
\usepackage{subfiles}
\usepackage{readarray}
\usepackage{xcolor}
\usepackage{import}
\usepackage{hhline}
\usepackage{bbm}

\newcommand{\softmax}{\operatornamewithlimits{softmax}}

\aclfinalcopy 

\title{Enhanced Universal Dependency Parsing with Second-Order Inference and Mixture of Training Data}

\author{Xinyu Wang$^{\diamond}$, Yong Jiang$^{\dagger}$,  Kewei Tu$^{\diamond}$\\
 $^\diamond$School of Information Science and Technology, ShanghaiTech University \\
 Shanghai Engineering Research Center of Intelligent Vision and Imaging \\
 $^\dagger$DAMO Academy, Alibaba Group \\
  {\tt \{wangxy1,tukw\}@shanghaitech.edu.cn} \\
  {\tt yongjiang.jy@alibaba-inc.com} \\
 }

\date{}

\begin{document}
\maketitle
\begin{abstract}
This paper presents the system used in our submission to the \textit{IWPT 2020 Shared Task}. Our system is a graph-based parser with second-order inference. For the low-resource Tamil corpus, we specially mixed the training data of Tamil with other languages and significantly improved the performance of Tamil. Due to our misunderstanding of the submission requirements, we submitted graphs that are not connected, which makes our system only rank \textbf{6th} over 10 teams. However, after we fixed this problem, our system is 0.6 ELAS higher than the team that ranked \textbf{1st} in the official results.
\end{abstract}

\section{Introduction}
Based on the Universal Dependencies (UD) \cite{nivre-etal-2016-universal}, the Enhanced Universal Dependencies (EUD) \cite{EUDparsingST:2020}\footnote{\url{https://universaldependencies.org/u/overview/enhanced-syntax.html}} are non-tree graphs with reentrancies, cycles, and empty nodes to deal with the problem that purely rooted trees cannot adequately represent grammatical relations. We found that we can reduce parsing such a graph to parsing bi-lexical structures like semantic dependency parsing (SDP) \cite{oepen2015semeval} by reducing reentrancies and empty nodes into new labels. \cite{wang-etal-2019-second} is a state-of-the-art approach for the semantic dependency parsing tasks that use second-order inference methods with Mean-Field Variational Inference. We adopt their approach for decoding and encode the sentences with strong pre-trained token representations: XLMR \cite{conneau2019unsupervised}, Flair \cite{akbik-etal-2018-contextual} and FastText \cite{bojanowski2017enriching}. Among the datasets, the Tamil language only contains 400 labeled sentences for training, which makes the performance of the model for Tamil low. To further improve the performance for the low resource language, we propose a new approach that we train the Tamil model with a mixture of datasets with Tamil and a rich resource language. Empirical results show that such an approach can improve 2.44 ELAS on the test set of Tamil. Due to our misconceptions on the submission format, we submitted invalid unconnected graphs to the submission site. Thanks to the help of the organizers, they fixed these graphs with simple scripts, and our system is ranked \textbf{6th} over 10 teams in the official results. However, the submitted graphs can be easily connected if we apply tree algorithms in the decoding. In the post-evaluation, we submitted our system outputs again and found that our system is 0.56 ELAS higher than the team ranked \textbf{1st} in the official results. 

\section{System Description}
\subsection{Data Pre-processing}
There are two features in the EUD graphs that do not appear in SDP graphs. One is the reentrancies of the same head and dependent on different labels. We combined these arcs into one and concatenate the labels of these arcs with a symbol `+` representing the combination of two arcs. In the post-processing, we split arcs with the `+` symbol in the corresponding labels into multiple arcs. Another one is the empty nodes that are introduced in the shared task (for example, nodes with id 1.1). We used the official script to collapse graphs through reducing such empty nodes into non-empty nodes and introducing new dependency labels\footnote{For more details, please refer to \url{https://universaldependencies.org/iwpt20/task_and_evaluation.html}.}. In the post-process, we add empty nodes according to the dependency labels. As the official evaluation only score the collapsed graphs, such a process does not impact the system performance.

\subsection{Approach}
We follow the approach of \citet{wang-etal-2019-second}\footnote{\url{https://github.com/wangxinyu0922/Second_Order_SDP}} to build our system which uses the second-order inference algorithm for the arc predictions. Given a sentence with $n$ words $\mathbf{w}=[w_1,w_2,...,w_n]$, we feed a three-layer BiLSTM with their corresponding token representations.
\begin{align*}
    \mathbf{R}&=\mathrm{BiLSTM}(\mathbf{E})
\end{align*}
where $\mathbf{E}=[\mathbf{e}_1,\dots,\mathbf{e}_n]$ is the concatenation of various embeddings of token (We use different combination of XLMR, Flair and FastText for each language as the token representation.) and $\mathbf{R}=[\mathbf{r}_1,\dots,\mathbf{r}_n]$ represents the output from the BiLSTM. For the arc predictions, we use the feed-forward network, Biaffine and Trilinear functions to encode unary potentials $\psi_{u}$ and binary potentials $\psi_{b}$: 
\begin{align*}
    \psi_{u}(w_i,w_j) &=\mathrm{FNN\_Biaffine}^{\mathrm{(arc)}}(\mathbf{r}_i,\mathbf{r}_j)\\
    \psi_{b}(w_i,w_j,w_k) &=\mathrm{FNN\_Trilinear}(\mathbf{r}_i,\mathbf{r}_j,\mathbf{r}_k)
\end{align*}
where $\mathrm{FNN\_Biaffine}$ and $\mathrm{FNN\_Trilinear}$ represent a combination of FNN and Biaffine/Trilinear functions. Then we feed these potentials into a Mean-Field Variational Inference network for the second-order inference. 
\begin{align*}
    P(\mathbf{Y}|\mathbf{w})=\mathrm{MFVI}(\psi_{u},\psi_{p})
\end{align*}
where $P(\mathbf{Y}|\mathbf{w})$ is a probability matrix representing the probabilities of all potential arcs. We first use tree algorithms like the Eisner's \cite{eisner2000bilexical} or MST \cite{mcdonald2005online} algorithms to ensure the connectivity of the graph. Then we additionally add arcs for the positions that $P(\mathbf{Y}|\mathbf{w})>0.5$.
For the label predictions, we use the $\mathrm{FNN\_Biaffine}$ to score the labels for each potential arc.
\begin{align*}
    \mathbf{s}_{ij}^{\mathrm{(label)}}=\mathrm{FNN\_Biaffine}^{\mathrm{(label)}}(\mathbf{r}_i,\mathbf{r}_j)\\
    P^{\mathrm{(label)}}(y_{ij}|\mathbf{w})=\softmax(\mathbf{s}_{ij}^{\mathrm{(label)}})
\end{align*}
We select the label with the highest score of each potential arc.

To train the system, we follow the approach of \citet{wang-etal-2019-second} with the cross entropy loss:
\begin{align*}
\mathcal{L}^{\mathrm{(arc)}} (\theta) &= -\sum_{i,j} \log(P_\theta (y_{ij}^{\star\mathrm{(arc)}}|\mathbf{w}))\\
\mathcal{L}^{\mathrm{(label)}} (\theta) &= -\sum_{i,j}\mathbbm{1}(y_{ij}^{\star\mathrm{(arc)}}) \log(P_\theta (y_{ij}^{\star\mathrm{(label)}}|\mathbf{w}))
\end{align*}
where $\theta$ is the parameters of our system, $\mathbbm{1}(y_{ij}^{\star\mathrm{(arc)}})$ denotes the indicator function and equals 1 when edge $(i,j)$ exists in the gold parse and 0 otherwise, and $i,j$ ranges over all the tokens $\mathbf{w}$ in the sentence.
The two losses are combined by a weighted average.
\begin{equation*}
    \mathcal{L}=\lambda\mathcal{L}^{(label)}+(1-\lambda)\mathcal{L}^{(arc)}
\end{equation*}

\subsection{Mixture of Datasets for Tamil Parser Training}
Tamil dataset has the fewest training and development sentences over all languages, which contains 400 sentences for training and 80 sentences for development. Therefore we believe that Tamil parser can be easily improved if we use more training data. With the emergence of multilingual contextual embeddings like multilingual BERT \cite{devlin-etal-2019-bert} and XLMR, training a unified multilingual model with high performances over all languages becomes possible through mixing the training data of multiple languages. However, it does not apply to the shared task as the label set of EUD is distinct in different languages. The arc annotations in the dataset are still helpful for training the Tamil parser. Thus we removed the label annotations in the dataset of other languages so that the label loss of these data cannot be back-propagated. Then we mixed one of the languages with the fully annotated Tamil dataset. To solve the problem of data imbalance in the mixture of the dataset in training, we upsampled the Tamil training set to keep the same data size as that of the other language.

\section{Settings and Results}
\subsection{Experimental Settings}
In training, we split the official development set into halves as the development set and test set. We used the development set to select the model based on labeled F1 score which is the metric used in the SDP task and it evaluates the accuracy of predicted labeled arcs. We used the test set to choose the best model architecture. We use a batch size of 2000 tokens with the Adam \cite{kingma2014adam} optimizer. The hyper-parameters of our system are shown in Table \ref{tab:hyper_both}, which are mostly adopted from previous work on dependency parsing. We only use the tokenized words as the model input. For the Tamil Parser, we tried English or Czech datasets to mix with the Tamil dataset. For most of languages, we used freezed XLMR embedding only as we found that the Flair embeddings and FastText embeddings were not helpful for the task except Tamil. We used a concatenate of XLMR, Flair and FastText embeddings for Tamil parser training. For the sentence and word segmentation, we used Stanza \cite{qi2020stanza} models that were trained on treebank with the largest training set for all languages except Lithuanian, because the model trained on the Lithuanian-HSE treebank has an extremely low segmentation performance compared with the model trained on Lithuanian-ALKSNIS. 
\begin{table}[t!]
\begin{center}
\begin{tabular}{lr}
\hline \hline
\textbf{Hidden Layer} & \textbf{Hidden Sizes}\\ \hline
BiLSTM LSTM & 3*400 \\
Unary Arc/Label & 500\\
Binary Arc & 150\\
Embedding/LSTM Dropouts & 33\%\\
Loss Interpolation ($\lambda$)& 0.10\\
Adam $\beta_1$ & 0.9\\
Adam $\beta_2$ & 0.9\\
Learning rate & $2e^{-3}$\\
LR decay & 0.5\\
\hline \hline
\end{tabular}
\end{center}
\caption{Hyper-parameters for our system. }
\label{tab:hyper_both}
\end{table}

\begin{table*}[ht!]
\small
\centering
\setlength\tabcolsep{3pt}
\begin{tabular}{l||c|c|c|c|c|c|c|c|c|c|c|c|c|c|c|c|c|c}
\hline\hline
Team Name           & ar   & bg   & cs   & nl   & en   & et   & fi   & fr   & it   & lv   & lt   & pl   & ru   & sl   & sv   & ta   & uk   & Avg. \\
\hline
&\multicolumn{17}{c}{Official}\\
\hline
RobertNLP           & 0.0  & 0.0  & 0.0  & 0.0  & \textbf{88.9} & 0.0  & 0.0  & 0.0  & 0.0  & 0.0  & 0.0  & 0.0  & 0.0  & 0.0  & 0.0  & 0.0  & 0.0  & 5.2  \\
Koebsala            & 60.8 & 68.9 & 61.1 & 62.9 & 65.4 & 59.1 & 67.5 & 67.9 & 69.1 & 64.8 & 56.3 & 61.3 & 64.2 & 64.1 & 64.5 & 47.4 & 64.2 & 62.9 \\
ADAPT               & 57.2 & 77.3 & 66.4 & 67.7 & 70.4 & 61.1 & 72.4 & 74.7 & 72.0 & 72.4 & 58.4 & 65.9 & 75.3 & 68.4 & 68.4 & 48.5 & 66.4 & 67.2 \\
clasp               & 51.3 & 84.9 & 67.1 & 78.9 & 82.9 & 60.4 & 66.0 & 72.8 & 87.1 & 66.0 & 52.6 & 71.2 & 70.4 & 65.2 & 71.4 & 42.2 & 63.2 & 67.9 \\
Ours                & 63.4 & 78.7 & 75.4 & 70.9 & 72.3 & 74.9 & 76.0 & 77.0 & 73.1 & 77.8 & 66.9 & 71.0 & 78.3 & 73.1 & 69.6 & 48.2 & 73.0 & 71.7 \\
Unipi               & 57.8 & 84.9 & 76.0 & 77.6 & 84.0 & 57.2 & 72.1 & 78.9 & 89.1 & 68.2 & 61.1 & 70.6 & 76.9 & 81.4 & 78.7 & 48.5 & 73.9 & 72.8 \\
FASTPARSE           & 66.9 & 84.9 & 77.2 & 77.4 & 78.5 & 74.1 & 75.7 & 77.8 & 84.8 & 75.6 & 61.4 & 74.5 & 80.4 & 73.5 & 75.2 & 47.0 & 74.0 & 74.0 \\
EmoryNLP            & 67.3 & 88.2 & 85.5 & 80.7 & 85.3 & 81.4 & 83.0 & \textbf{86.2} & 88.5 & 79.2 & 66.1 & 82.4 & 88.6 & 82.7 & 78.2 & 54.3 & 79.7 & 79.8 \\
OrangeDeskin        & 71.0 & 89.4 & 87.0 & 85.1 & 85.2 & 81.0 & 86.2 & 83.6 & 90.8 & 82.1 & 75.9 & 80.4 & 89.8 & 84.4 & 83.3 & \textbf{64.2} & 84.6 & 82.6 \\
TurkuNLP            & \textbf{77.8} & 90.7 & 87.5 & 84.7 & 87.2 & 84.5 & \textbf{89.5} & 85.9 & 91.5 & 84.9 & 77.6 & \textbf{84.6} & 90.7 & \textbf{88.6} & \textbf{85.6} & 57.8 & 87.2 & 84.5 \\
\hline
&\multicolumn{17}{c}{Post-Evaluation}\\
\hline
Ours+en+MST    & 77.7 & 91.5 & 90.1 & 86.2 & 87.1 & 86.0 & 89.0 & 85.3 & 91.5 & 87.6 & 78.9 & 84.0 & 92.3 & 87.6 & 84.7 & 56.7 & 88.0 & 85.0 \\
Ours+cs+Eis & 77.8 & 91.1 & 89.5 & 86.3 & 87.2 & 85.7 & 88.5 & 85.3 & \textbf{91.5} & 87.3 & 78.6 & 83.7 & 92.3 & 87.1 & 84.8 & 58.4 & 88.0 & 84.9 \\
Ours+cs+MST      & 77.7 & \textbf{91.5} & \textbf{90.1} & \textbf{86.2} & 87.1 & \textbf{86.0} & 89.0 & 85.3 & 91.5 & \textbf{87.6} & \textbf{78.9} & 84.0 & \textbf{92.3} & 87.6 & 84.7 & 58.4 & \textbf{88.0} & \textbf{85.1}\\
\hline
\hline
\end{tabular}
\caption{Official evaluations of all systems and post-evaluations of our team in ELAS. We use the ISO 639-1 language code to represent each language. MST and Eis means the MST and Eisner's algorithm that we used for decoding. ``en'' and ``cs'' represents which dataset we mixed with the Tamil dataset for training the Tamil parser. Note that `Ours+en+MST` represent the parsed results of parsers that we used in the Official submission.}
\label{tab:results}
\end{table*}

\begin{table*}[ht!]
\small
\centering

\begin{tabular}{l||c|c|c|c|c|c|c|c|c}
\hline\hline
Approach                      & ar    & bg    & cs    & nl    & en    & et    & fi    & fr    & it    \\
\hline\hline
XLMR+Flair+FastText+1st-Order & 81.66 & 89.29 & 91.04 & 92.55 & 89.74 & 88.33 & 89.40 & 90.64 & 91.94 \\
XLMR+Flair+FastText+2nd-Order & 81.98 & 89.43 & \textbf{91.39} & \textbf{92.68} & 89.58 & \textbf{88.69} & 89.54 & 91.08 & \textbf{91.98} \\
XLMR+1st-Order                & 82.02 & 90.15 & 90.80 & 92.43 & 90.05 & 88.13 & 89.51 & 91.14 & 91.96 \\
XLMR+2nd-Order                & \textbf{82.42} & \textbf{90.37} & 91.21 & 92.66 & \textbf{90.26} & 88.60 & 90.35 & \textbf{91.69} & \textbf{91.98} \\
\hline
                              & lv    & lt    & pl    & ru    & sk    & sv    & ta    & uk    & Avg.  \\
\hline\hline
XLMR+Flair+FastText+1st-Order & 88.21 & 80.21 & 86.91 & 92.88 & 87.28 & 85.52 & 66.17 & 88.26 & 89.40 \\
XLMR+Flair+FastText+2nd-Order & 88.59 & 81.25 & 86.46 & \textbf{93.28} & 87.18 & 85.63 & \textbf{68.76} & 88.04 & 89.59 \\
XLMR+1st-Order                & 89.62 & 81.92 & 85.73 & 92.86 & 88.48 & 86.36 & 63.28 & 88.96 & 89.57 \\
XLMR+2nd-Order                & \textbf{89.97} & \textbf{83.24} & \textbf{87.49} & 93.21 & \textbf{89.07} & \textbf{86.85} & 64.84 & \textbf{89.99} & \textbf{89.95}\\
\hline\hline
\end{tabular}
\caption{A comparison of different word embedding concatenation and first-order and second-order inference approaches on the development set split by ourselves. We report Labeled F1 score (LF1) here.}
\label{tab:comparison}
\end{table*}

\begin{table*}[ht!]
\small
\centering
\setlength\tabcolsep{3.5pt}
\begin{tabular}{l||cccccccc}
\hline\hline
Graph          & ar-PADT      & bg-BTB  & cs-FicTree & cs-CAC & cs-PDT  & cs-PUD & nl-Alpino    & nl-LassySmall \\
\hline
Non-Connected & 77.74        & \textbf{91.50}   & \textbf{90.60}      & 90.55  & \textbf{90.65}   & \textbf{84.26}  & \textbf{90.11}        & 82.55         \\
MST            & 77.73        & 91.48   & 90.51      & \textbf{90.59}  & 90.63   & 84.25  & 90.09        & 82.51         \\
Eisner's            & \textbf{77.75}        & 91.07   & 89.85      & 90.02  & 90.05   & 83.70  & 89.69        & \textbf{83.10}         \\
\hline\hline
Graph          & en-EWT       & en-PUD  & et-EDT     & et-EWT & fi-TDT  & fi-PUD & fr-Sequoia   & fr-FQB        \\
\hline
Non-Connected & 86.33        & \textbf{88.05}   & \textbf{87.36}      & \textbf{79.62}  &    \textbf{90.00} & \textbf{87.52}  & 89.67        & 84.11         \\
MST            & 86.30        & \textbf{88.05}   & 87.34      & 79.61  & 89.97   & \textbf{87.52}  & 89.66        & 84.09         \\
Eisner's            & \textbf{86.40}        & 88.04   & 87.07      & 79.42  & 89.44   & 86.97  & \textbf{89.73}        & \textbf{84.12}         \\
\hline\hline
Graph          & it-ISDT      & lv-LVTB & lt-ALKSNIS & pl-LFG & pl-PDB  & pl-PUD & ru-SynTagRus & sl-SNK        \\
\hline
Non-Connected & 91.50        & \textbf{87.69}   & 78.97      & \textbf{87.65}  & \textbf{83.23}   & \textbf{82.96}  & \textbf{92.62}        & \textbf{87.56}         \\
MST            & 91.49        & 87.64   & \textbf{78.94}      & \textbf{87.65}  & 83.21   & 82.95  & 92.31        & 87.55         \\
Eisner's            & \textbf{91.52}        & 87.29   & 78.63      & 87.59  & 82.90   & 82.57  & 92.31        & 87.14         \\
\hline\hline
Graph          & sv-Talbanken & sv-PUD  & ta-TTB     & uk-IU  & Average &        &              &               \\
\hline
Non-Connected &    88.35     & \textbf{80.88}   & 56.51      & 88.00  & \textbf{85.59}   &        &              &               \\
MST            & 88.33        & 80.87   & 56.56      & \textbf{88.02}  & 85.57   &        &              &               \\
Eisner's            & \textbf{88.37}        & 80.87   & \textbf{56.71}      & \textbf{88.02}  & 85.37   &        &              &              \\
\hline\hline
\end{tabular}
\caption{A performance comparison in ELAS between non-connected graphs and connected graphs over each treebank on the official test sets.}
\label{tab:tb_connected}
\end{table*}

\begin{table*}[ht!]
\centering
\begin{tabular}{l||c|c|c|c|c|c|c|c|c}
\hline\hline
Team Name   & ar    & bg    & cs    & nl    & en    & et    & fi    & fr    & it    \\
\hline
TurkuNLP    & \textbf{77.82} & 90.73 & 87.51 & 84.73 & \textbf{87.15} & 84.54 & \textbf{89.49} & \textbf{85.90} & \textbf{91.54} \\
Ours+en+MST & 77.74 & 91.48 & 90.09 & 86.19 & 87.10 & 85.97 & 88.99 & 85.28 & 91.49 \\
Ours+en     & 77.75 & \textbf{91.50} & \textbf{90.11} & \textbf{86.22} & 87.12 & \textbf{85.99} & 89.01 & 85.29 & 91.50 \\
\hline\hline
Team Name   & lv    & lt    & pl    & ru    & sl    & sv    & ta    & uk    & Avg.  \\
\hline
TurkuNLP    & 84.94 & 77.64 & \textbf{84.64} & 90.69 & \textbf{88.56} & \textbf{85.64} & \textbf{57.83} & 87.22 & 84.50 \\
Ours+en+MST & 87.64 & 78.94 & 84.00 & 92.31 & 87.55 & 84.74 & 56.71 & \textbf{88.02} & 84.96 \\
Ours+en     & \textbf{87.69} & \textbf{78.97} & 84.01 & \textbf{92.62} & 87.56 & 84.75 & 56.52 & 88.00 & \textbf{84.98} \\
\hline\hline
\end{tabular}
\caption{A performance comparison in ELAS between non-connected graphs, connected graphs with the MST algorithm and the best system in the official results over each language. Ours+en represents our official submission and evaluated with official evaluation script.}
\label{tab:connected}
\end{table*}

\begin{table}[t!]
\setlength\tabcolsep{5pt}
\begin{center}
\begin{tabular}{l|cc}
\hline \hline
\textbf{Combination} & \textbf{\# Training Sentences} & \textbf{ELAS}\\ 
\hline
Tamil & 400 & 55.39\\
English+Tamil & 12543+12400 & 56.56\\
Czech+Tamil & 102131+102000 & \textbf{58.44}\\
\hline \hline
\end{tabular}
\end{center}
\caption{A comparison between different dataset combinations for the Tamil parser training. The 12400 and 102000 in the \textbf{\# Training Sentences} column represents the upsampled value of 400 labeled sentences in Tamil dataset.}
\label{tab:dataset_comp}
\end{table}

\subsection{Main Results}
Table \ref{tab:results} shows the results of official evaluations of all teams, as well as the post-evaluations of our system. In the Official submission, we trained the Tamil Parser with a mixture of English and Tamil datasets (`Ours+en+MST` in the table), and in the post-evaluation, we also tried a mixture of Czech and Tamil datasets (`Ours+cs+MST` in the table) because the Czech dataset contains the largest training data over all languages. In the official results, our system was fixed by the organizers through their simple scripts for the connectivity of graphs, which significantly reduced our system performance. In the post-evaluation, we fixed this issue with MST or Eisner's algorithm and showed that our system performs 0.6 ELAS higher than the best team. For the Tamil parser, mixing the Tamil dataset with the Czech dataset performs 1.7 ELAS better than mixing with the English dataset, which shows that a larger dataset gives better results than the smaller one. Our system with the MST algorithm is 0.2 ELAS stronger than the system with Eisner's algorithm, which shows that the non-projective tree algorithm (MST) is better than the projective tree algorithm (Eisner's) for the EUD task. We built our codes based on PyTorch \cite{paszke2019pytorch}, and ran our experiments on a single Tesla V100 GPU.

\subsection{Comparison of First-Order and Second-Order Inference}
Table \ref{tab:comparison} shows a performance comparison between two kinds of embedding choices, XLMR+Flair+FastText and XLMR, and first-order and second-order inference. The results show that second-order inference is stronger than first-order inference in all languages, and embeddings with XLMR embedding only usually perform better than XLMR+Flair+FastText embeddings. However, the Flair+FastText embedding is helpful for Tamil. Therefore we use XLMR+Flair+FastText embeddings for training the Tamil parser while we use XLMR embedding only for other languages.

\subsection{Performance Comparison between Connected Graphs and Non-Connected Graphs}
Before the deadline of the shared task, the submission site showed the scores of each treebank separately even the submission graphs were not connected, which unfortunately made us believe that the non-connected graphs are also acceptable for the task. In fact, these graphs are not acceptable and the organizers fixed the issue with some simple scripts, and this results in a significant reduction in the final scores. In section 3.2, we show that appending a tree-parsing algorithm to our system produces connected graphs with high scores. Here we also evaluate the non-connected graphs produced by our original system. We think evaluating non-connected graphs is informative for two reasons. The first is that these results help to understand how different the connected graphs and non-connected graphs performs. The second is that in practice, non-connected graphs can be predicted with a relatively faster speed as the MST and Eisner's algorithms are slow while we can get the non-connected graphs through argmax operations. We compare the performance of connected and non-connected graphs for each treebank and each language in Table \ref{tab:tb_connected} and \ref{tab:connected}. The results show that the non-connected graphs perform slightly better than graphs with the tree algorithms. Therefore generating non-connected trees are more practical in practice if there are no such constraints.

\subsection{Analysis of Mixture of Training Data}
For a more in-depth comparison of how the combination of different language datasets affects the performance of the Tamil Parser, Table \ref{tab:dataset_comp} shows that more training data significantly improve the performance of the parser. We leave for future work other language combinations as well as similar studies of other parsers.

\section{Conclusion}
Our system is a parser with strong contextual embeddings and second-order inference. For the low-resource language, we propose to train the model with a mixture of datasets. Empirical results show that the second-order inference is stronger than the first-order one, and mixing data improves the performance of parser significantly for the low-resource language. After we fix the graph connectivity issue, our system outperforms the system ranked \textbf{1st} by 0.56 ELAS in the official results. We also show that the non-connected graphs are practically useful for its higher performance and faster speed. Our code is available at \url{https://github.com/Alibaba-NLP/MultilangStructureKD}.

\bibliography{anthology,acl2020}
\bibliographystyle{acl_natbib}

\end{document}